\documentclass{article}
\usepackage{silence}
\usepackage[preprint]{neurips_2025}

\usepackage[utf8]{inputenc}
\usepackage[T1]{fontenc}
\usepackage{hyperref}
\usepackage{url}
\usepackage{booktabs}
\usepackage{amsmath,amssymb,amsthm}
\usepackage{graphicx}
\usepackage{wrapfig}
\usepackage{multirow}
\usepackage{adjustbox}
\usepackage{microtype}
\usepackage{color}
\usepackage{algorithmic}
\usepackage{wrapfig}
\usepackage{float}
\usepackage{enumitem}
\usepackage{parskip} % Adds space between paragraphs, good for readability

\title{Twist and Compute: The Cost of Pose in 3D Generative Diffusion}

\author{Kyle Fogarty
 \\
University of Cambridge \\
\texttt{ktf25@cam.ac.uk}
\And
Jack Foster\\
University of Cambridge\\
\texttt{jwf40@cam.ac.uk}
\And
 Boqiao Zhang\\
 University of Cambridge\\
 \texttt{bz317@cam.ac.uk}
\And
Jing Yang\\
University of Cambridge\\
\texttt{jy496@cam.ac.uk}
\And
Cengiz Öztireli\\
University of Cambridge\\
\texttt{aco41@cam.ac.uk}
}

\begin{document}
\maketitle

% -------------------- ABSTRACT --------------------
\begin{abstract}
Despite their impressive results, large-scale image-to-3D generative models remain opaque in their inductive biases.
We identify a significant limitation in image-conditioned 3D generative models: a strong canonical view bias.
Through controlled experiments using simple 2D rotations, we show that the state-of-the-art Hunyuan3D 2.0 model can struggle to generalize across viewpoints, with performance degrading under rotated inputs.
%NEED A CONNECTOR SENTENCE HERE>
We show that this failure can be mitigated by a lightweight CNN that detects and corrects input orientation, restoring model performance without modifying the generative backbone. 
 Our findings raise an important open question: Is scale enough, or should we pursue modular, symmetry-aware designs?
\end{abstract}

% -------------------- 1. INTRODUCTION --------------------
\section{Introduction}
\begin{wrapfigure}{r}{0.41\textwidth} % 'r' = right side, width of figure box
  \centering
  \includegraphics[width=0.39\textwidth]{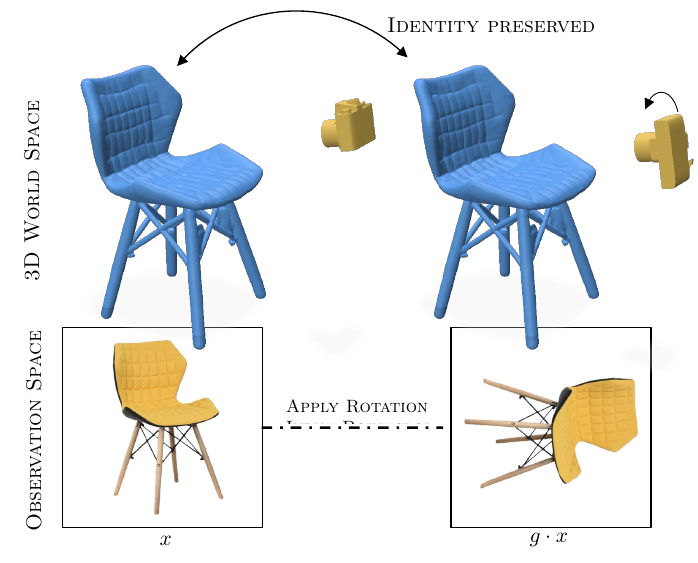}
  \caption{Equivariance between world and observation spaces preserves identity.}
  \label{fig:wrap}
\end{wrapfigure}
The expressivity of generative models, particularly diffusion architectures, has led to unprecedented success in 3D content creation from single images. These models promise to learn the complex distribution from a mixture of curated 3D training data and image foundation models. However, this success raises a foundational question: What inductive biases are actually learned during training, and how do they impact generalization? In the physical world, objects retain their identity despite changes in position and orientation. However, when these objects are projected into an observation space, such as an image, transformations result in structured variations in appearance. A model that genuinely understands 3D structure should therefore maintain object identity across these transformations, exhibiting equivariance between the 3D world space and its representation in the observation space.

Established works argue that symmetry should be encoded rather than merely discovered: group-equivariant CNNs enforce rotation/reflection structure in 2D \cite{cohen2016group}, and SE(3)-equivariant networks extend this to 3D representations \cite{thomas2018tensorfield, fuchs2020se3transformers}. In contrast, popular image-to-3D pipelines often inherit dataset view biases (e.g., canonical front views), which can encourage shortcut solutions over true geometry \cite{chan2022eg3d,liu2023zero123}. Recent efforts try to counter this bias through two main strategies: fine-tuning models on orientation-aligned data to directly produce canonical outputs \cite{orientationaligned2025}, or normalizing the input pose before reconstruction. The viability of the latter is supported by foundational work showing that simple networks can effectively predict 2D image rotation \cite{gidaris2018unsupervised}.
% Recent efforts try to counter this with lightweight canonicalization or rotation-prediction modules that normalize pose before reconstruction \cite{gidaris2018unsupervised,orientationaligned2025}.\\

% {\color{red} @Kyle: Todo, reformat Fig.~\ref{fig:overview_figre}}.
\begin{figure}[]
    \centering
    \includegraphics[width=\linewidth]{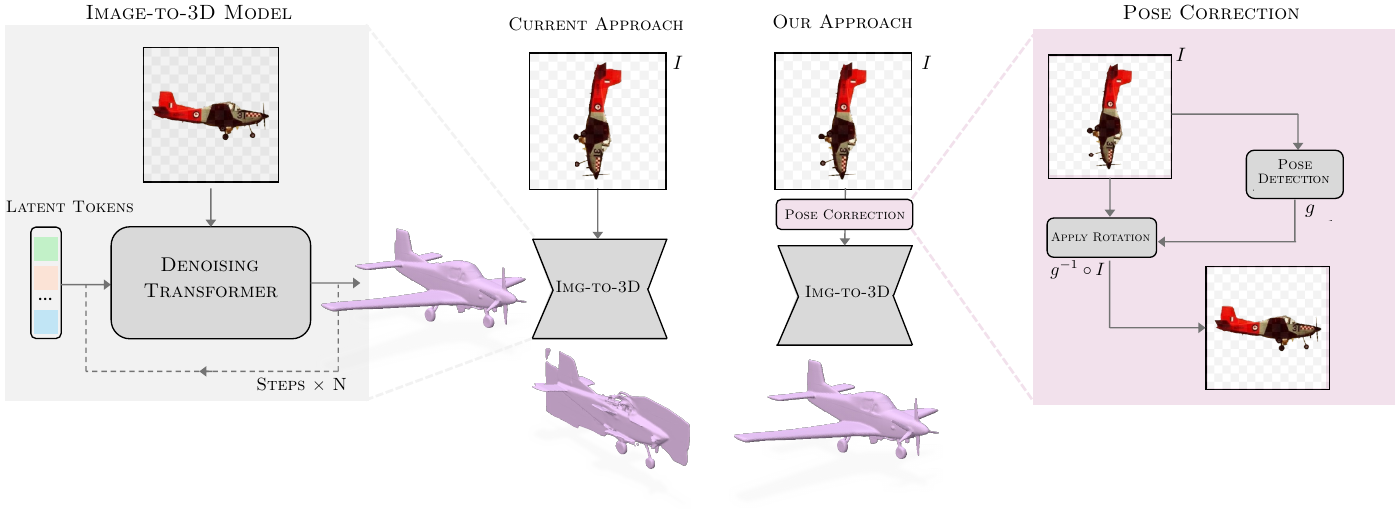}
    \caption{Rotating the input image breaks the image-to-3D pipeline (canonical-view bias), while a lightweight CNN predicts the rotation and applies its inverse to re-canonicalize the image before 3D generation, restoring performance. Note that the degraded 3D mesh is rotated to the canonical frame for better visualization.}
    \label{fig:overview_figre}
\end{figure}

We position our study within this landscape, using in-plane rotations of the input image as a controlled probe of whether
the Hunyuan3D 2.0 image-to-3D pipeline \cite{zhao2025hunyuan3d}, a current state-of-the-art model, exhibits genuine equivariance or  instead relies upon on canonical views. A model that internalizes 3D structure should be equivariant to such transformations, maintaining object identity as the observation rotates. However, the prevalence of canonical orientations in large-scale image corpora may incentivize shortcut learning, yielding a preference for canonical views rather than robust 3D geometry. We refer to this as \textit{canonical-view bias}, which limits generalization to arbitrarily oriented inputs.
% In this paper, we investigate a simple subset of this challenge: rotational symmetry in the 2D input image. We hypothesize that due to the strong canonical orientation of objects in typical web-scale datasets (e.g., cars are rarely shown upside down), these massive models learn a \textit{shortcut}, developing a strong bias for canonical views rather than a true understanding of 3D geometry. This canonical view bias is a failure of generalization, limiting the model's utility on real-world inputs which may not be perfectly aligned.

% \paragraph{Contributions} We (i) quantify a strong orientation bias in the Hunyuan3D image-to-3D model on canonical-pose categories (airplanes/chairs/cars), (ii) show the resulting catastrophic 3D failures, and (iii) restore performance with a tiny CNN pre-processor that canonicalizes image orientation directly.
% \paragraph{Contributions} (i) We empirically reveal a strong orientation (canonical view) bias in the Hunyuan3D image-to-3D generative model, specifically for categories with well-defined canonical poses such as airplanes, chairs, and cars. (ii) We demonstrate that this bias leads to failures in 3D generation when inputs are rotated away from canonical orientations. (iii) We show that a simple, lightweight CNN pre-processing module, trained to detect and correct image orientation, can restore generation quality without modifying the underlying generative model.

\paragraph{Contributions}
This work investigates the impact of viewpoint biases in image-to-3D generation.
(i) We empirically identify a strong canonical view bias in the Hunyuan3D generative model, particularly for object categories with well-defined orientations, such as airplanes, chairs, and cars.
(ii) We demonstrate that this bias can significantly degrade 3D generation quality when input images are rotated away from their canonical poses.
(iii) We show that a lightweight CNN-based pre-processing module, trained to detect and correct image orientation, can effectively restore generation quality, without requiring any changes to the generative model itself.
% Our contributions are threefold:
% \begin{enumerate}
%     \item We provide a rigorous empirical analysis quantifying a severe orientation bias in the Hunyuan image-to-3D model using a curated dataset of objects with clear canonical poses (airplanes, chairs, and cars).
%     \item We show, through striking qualitative examples, how this bias leads to catastrophic failures in the generated 3D geometry.
%     \item We propose and validate a remarkably simple solution: a tiny CNN pre-processor that canonicalizes image orientation, restoring the performance of the multi-billion parameter generative model.
% \end{enumerate}
% This work directly addresses the workshop's call to understand "learnability" and "inductive biases," questioning whether fundamental symmetries should be learned implicitly or "hardwired" into the modeling pipeline for more robust and efficient systems.

% \section{Background}

% Our analysis is based on the Hunyuan img23D model, and so we provide a short introduction to the model. For full details, we refer the read to the technical report.

%%%%%%%%%%%%%%%%%%%%%%%%%%%%%%%%%%%%%%%%%%%%%%%%%%%%%%%%%%%%%%%%
% SECTION 2: METHODOLOGY
%%%%%%%%%%%%%%%%%%%%%%%%%%%%%%%%%%%%%%%%%%%%%%%%%%%%%%%%%%%%%%%%
\section{Methodology}

\subsection{Image-to-3D Generative Model}

Our investigation focuses on Hunyuan3D \cite{zhao2025hunyuan3d}, a state-of-the-art flow-matching architecture for single-image 3D generation. The model adopts a decoupled pipeline, separating 3D representation learning from the generative process. To obtain the 3D representation, high-level semantic features are extracted from the input image using a frozen DINOv2-based encoder \cite{oquab2024dinov2learningrobustvisual}. These features are projected into a latent space as vector sets \cite{zhang20233dshape2vecset}, which serve as implicit representations of complex 3D shapes. The generative component is a flow-matching diffusion transformer trained in this latent space to predict object token sequences from the input image. These token sequences are decoded into Signed Distance Functions (SDFs), which are subsequently converted into triangle meshes via iso-surfacing \cite{lorensen1998marching}.

This architecture implies that the model’s synthesis quality is bottlenecked by the feature fidelity of its 2D encoder. If the DINOv2 encoder exhibits a bias toward canonical object views, it may produce distorted feature representations when processing rotated inputs.

% {\color{red} @Steven: I think it would be useful to add a bit of a technical introduction to Hunyuan3D here; we dont need to cover everything but would be useful to understand how we go from images (Dino features etc) to the \texttt{setvec} representation and then produce the mesh). Our aim here might be to cast a light on underlying mechanisms why Hunyuan fails in this task.}

\newpage
\subsection{Probing for Canonical View Bias}

\paragraph{Curated Dataset.}  
To investigate the model's sensitivity to input orientation, we constructed a targeted evaluation dataset comprising three object categories with well-established canonical poses: airplanes, chairs, and cars. Images were sourced from publicly available repositories and manually curated to ensure the presence of a single, dominant object in each image, minimal occlusion, and sufficient diversity across object instances.

\paragraph{Rotational Transformations.}  
To systematically evaluate robustness to viewpoint variation, we apply a set of in-plane 2D rotations to each image in the dataset. Specifically, each source image \( I \) is rotated by angles \( \theta \in \{0^\circ, 90^\circ, 180^\circ, 270^\circ\} \), producing a set of transformed inputs \( \{I'_{0}, I'_{90}, I'_{180}, I'_{270}\} \). Here, \( 0^\circ \) denotes the original, canonical orientation.

\paragraph{Evaluation Metric: Cross-Modal Similarity (ULIP).}  
To quantitatively assess the semantic fidelity of the generated 3D shapes, we employ the ULIP (Unified Language-Image Pre-training) score~\cite{xue2023ulip}, which enables direct comparison between modalities by embedding 2D images and 3D shapes into a shared semantic space. In our setup, each generated mesh is converted into a point cloud of 8{,}192 points, and 3D features are extracted using a pretrained Point-BERT variant integrated within the ULIP framework. The ULIP score is computed as the cosine similarity between the image and point cloud embeddings, with higher scores indicating stronger semantic alignment between the input image and the generated 3D shape.

\paragraph{Evaluation Procedure.}  
For each rotated image \( I'_{\theta} \), we use the Hunyuan3D model to generate a corresponding 3D shape \( M'_{\theta} \), and compute the associated ULIP score:
\[
\mathcal{S}_{\theta} = \text{ULIP}(I'_{\theta}, M'_{\theta}).
\]
By comparing scores across rotation angles, we assess the model’s robustness to input orientation and identify any performance degradation when the input deviates from the canonical view.

\subsection{A Lightweight CNN-based Orientation Corrector}
% {\color{red} @Steven: Would be good to expand this section a little more; we dont need to put all the information in-text but we can add a section in the appendix for example. The repo used is here: \texttt{https://github.com/duartebarbosadev/deep-image-orientation-detection}.}

% To mitigate orientation bias, we introduce a lightweight pre-processing module: a compact orientation classifier based on EfficientNetV2, equipped with a 4-way softmax output. Given an input image $I$, the classifier produces logits $\mathbf{z}(I) \in \mathbb{R}^4$, corresponding to rotation angles ${0^\circ, 90^\circ, 180^\circ, 270^\circ}$. The predicted rotation $r^\star$ is determined by $r^\star = \arg\max_k z_k(I)$, and the image is subsequently rotated to a canonical orientation. We use publicly available pretrained weights without fine-tuning. With only 20.3M parameters, negligible compared to our 2.8B parameter generative model, the classifier achieves 96.2\% accuracy in our tests.

To mitigate orientation bias, we introduce a lightweight pre-processing module: a compact orientation classifier based on EfficientNetV2, equipped with a 4-way softmax output. We provide more information on the orientator in Appendix \ref{appendix}. Given an input image \( I \), the classifier produces logits \( \mathbf{z}(I) \in \mathbb{R}^4 \), corresponding to discrete rotation angles \( \{0^\circ, 90^\circ, 180^\circ, 270^\circ\} \). The predicted rotation \( r^\star \) is obtained by
\[
r^\star = \arg\max_k z_k(I),
\]
after which the image is rotated back to a canonical orientation. We employ publicly available pretrained weights for the classifier without any additional fine-tuning. With only 20.3M parameters, negligible compared to the 2.8B parameters of the generative model, the classifier achieves 96.2\% accuracy in our evaluations.

%%%%%%%%%%%%%%%%%%%%%%%%%%%%%%%%%%%%%%%%%%%%%%%%%%%%%%%%%%%%%%%%
% SECTION 3: RESULTS AND ANALYSIS
%%%%%%%%%%%%%%%%%%%%%%%%%%%%%%%%%%%%%%%%%%%%%%%%%%%%%%%%%%%%%%%%
% \section{Results and Analysis}

% \paragraph{A Strong Canonical View Bias Leads to Generation Failure.}
% Our experiments confirm a severe performance degradation for non-canonical inputs. As shown in Figure \ref{fig:main_result}(a), the ULIP score is consistently highest for the canonical $0^\circ$ view and drops substantially for all other orientations across all object categories.
\section{Results and Analysis}
\label{sec:results}

% \subsection{Canonical-View Bias in Hunyuan3D~2.0}
% Across \emph{airplanes}, \emph{chairs}, and \emph{cars}, rotating the input image away from the canonical \(0^\circ\) view yields a marked decline in ULIP similarity. The \(0^\circ\) view consistently attains the highest score, while \(90^\circ\), \(180^\circ\), and \(270^\circ\) underperform, indicating sensitivity to input orientation rather than stable 3D understanding (Fig.~\ref{fig:main_result}). Qualitatively, non-canonical inputs produce systematic geometric errors: e.g., collapsed or sheared airplane wings and misaligned or duplicated chair legs, whereas the \(0^\circ\) view remains stable (Fig.~\ref{fig:placeholder}).
\subsection{Canonical-View Bias in Hunyuan3D~2.0}

\begin{figure}[h!]
    \centering
    \includegraphics[width = \textwidth]{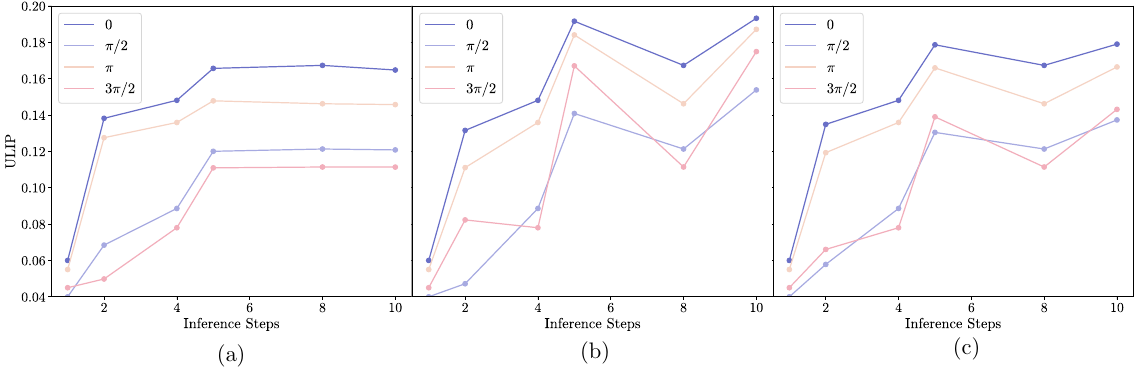}
    \caption{ULIP similarity (higher is better) versus diffusion inference steps for Hunyuan3D~2.0 under four in-plane input rotations $\{0^\circ, 90^\circ, 180^\circ, 270^\circ\}$, shown separately for (a) airplanes, (b) chairs, and (c) cars. Across all categories, the canonical $0^\circ$ view consistently achieves the highest ULIP, while non-canonical rotations suffer substantial drops.}
    \label{fig:main_result}
\end{figure}

We observe a consistent decline in ULIP similarity scores as the input image is rotated away from the canonical \( 0^\circ \) orientation across all evaluated categories: \emph{airplanes}, \emph{chairs}, and \emph{cars}. In each case, the \( 0^\circ \) view yields the highest similarity, while rotated views at \( 90^\circ \), \( 180^\circ \), and \( 270^\circ \) result in significantly lower scores, revealing a clear sensitivity to input orientation rather than robust 3D understanding (Fig.~\ref{fig:main_result}).

\begin{figure}
    \centering
    \includegraphics[width=\linewidth]{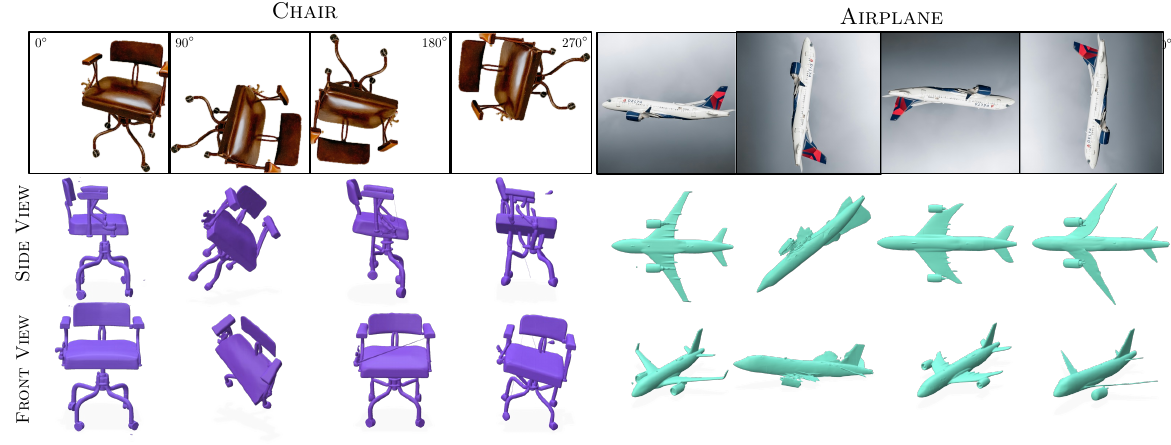}
    \caption{Qualitative effect of input rotation on Hunyuan3D 2.0. For each object, the input image is rotated by \{0°, 90°, 180°, 270°\}. To ensure comparability, all generated meshes are reoriented to a common camera before being rendered from canonical front and side views. Non-canonical inputs induce systematic geometric failures (e.g., collapsed airplane wings, misaligned/duplicated chair legs), whereas the 0° view remains stable, illustrating a strong canonical-view bias.}
    \label{fig:placeholder}
\end{figure}

This trend is further supported by qualitative analysis. Inputs at non-canonical angles often lead to systematic geometric failures; for example, collapsed or sheared airplane wings, and misaligned or duplicated chair legs, whereas outputs from canonical views remain structurally coherent and faithful to the object’s shape (Fig.~\ref{fig:placeholder}).

% \begin{figure}[h]
%   \centering
%   % TODO: Create your primary figure. This is the most important part of the paper.
%   % SUGGESTION:
%   % (a) A bar or line chart showing ULIP score vs. Rotation for each object category.
%   % (b) A grid of qualitative examples. Rows: Airplane, Chair, [Object 3].
%   %     Cols: Canonical Input, Rotated Input, Generated 3D from Rotated Input, Generated 3D after our Corrector.
%   \includegraphics[width=\linewidth]{placeholder.png}
%   \caption{\textbf{Left (a):} ULIP score as a function of input image rotation. Performance systematically degrades for non-canonical views. \textbf{Right (b):} Qualitative examples. A rotated input (Col 2) causes catastrophic failures in the generated 3D model (Col 3). Our simple CNN corrector restores the canonical view and enables high-quality generation (Col 4).}
%   \label{fig:main_result}
% \end{figure}

% \paragraph{The Effect of Inference Steps.}
% We also investigated whether increasing the number of diffusion inference steps could mitigate this bias. We found no consistent trend. While a higher number of steps occasionally led to minor improvements for non-canonical views, the effect was not reliable and did not close the significant performance gap. This suggests the orientation bias is a deeply ingrained feature of the model's learned data manifold, not an artifact of an incomplete generation process.

\paragraph{Effect of Inference Steps.}
We evaluated whether increasing the number of diffusion inference steps mitigates the orientation bias. Across categories and view angles, we observed no consistent trend: additional steps sometimes yielded small gains for non-canonical inputs, but the improvements were unstable and insufficient to close the gap to the \(0^\circ\) view. This suggests the bias is embedded in the learned representation rather than an artifact of under-converged sampling.

% \subsection{Lightweight Orientation Corrector}
% To mitigate this sensitivity, we apply a compact CNN-based orientation corrector to re-canonicalize inputs prior to generation. ULIP scores after correction closely match those of images originally at \(0^\circ\), without modifying the generative backbone (Table~\ref{tab:results}). These results support a simple, modular remedy for the observed canonical-view bias.

\subsection{Lightweight Orientation Corrector}

To address the model's sensitivity to input orientation, we introduce a compact CNN-based orientation corrector that re-canonicalizes input images prior to 3D generation. Importantly, this correction is applied as a pre-processing step and does not require any modification to the generative backbone. After correction, ULIP scores closely align with those obtained from images originally presented at the canonical \( 0^\circ \) view (Table~\ref{tab:results}), demonstrating the effectiveness of this lightweight intervention.

% \paragraph{Our Lightweight Corrector Recovers Performance.}
% The application of our CNN orientation corrector proves highly effective. As shown in Table \ref{tab:results}, the average ULIP score for images first passed through our corrector is nearly identical to the score of images that were already in the canonical orientation. This simple, low-parameter module successfully "plugs the hole" in the larger model's capability.

% \begin{table}[h]
%   \caption{ULIP scores across different conditions, averaged over all three object categories. Our lightweight CNN corrector restores performance to near-canonical levels.}
%   \label{tab:results}
%   \centering
%   \begin{tabular}{lc}
%     \toprule
%     Input Image Condition & Mean ULIP Score $\pm$ Std Dev \\
%     \midrule
%     Canonical View ($0^\circ$)        & $0.XX \pm 0.XX$ \\ % TODO: Fill with your data
%     Rotated View (Avg. over non-canonical) & $0.YY \pm 0.YY$ \\ % TODO: Fill with your data
%     Rotated View + CNN Corrector      & $0.ZZ \pm 0.ZZ$ \\ % TODO: Fill with your data
%     \bottomrule
%   \end{tabular}
% \end{table}
\begin{table}[hb!]
  \caption{ULIP scores (mean $\pm$ std) by category and input condition. Our lightweight CNN corrector restores performance close to the canonical view across all categories.}
  \label{tab:results}
  \centering
  \begin{tabular}{lccc}
    \toprule
    \multicolumn{1}{c}{Input Image Condition} &
    \multicolumn{1}{c}{Airplanes} &
    \multicolumn{1}{c}{Chairs} &
    \multicolumn{1}{c}{Cars} \\
    \midrule
    Canonical View ($0^\circ$)              & $0.166\,\pm\,0.047$ & $0.184\,\pm\,0.07$ & $0.176\,\pm\,0.07$ \\
    Rotated View (avg.\ non-canonical)      & $0.126\,\pm\,0.048$ & $0.155\,\pm\,0.06$ & $0.132\,\pm\,0.06$ \\
    Rotated View + CNN Corrector            & $0.166\,\pm\,0.047$ & $0.181\,\pm\,0.06$ & $0.170\,\pm\,0.07$ \\
    \bottomrule
  \end{tabular}
\end{table}

%%%%%%%%%%%%%%%%%%%%%%%%%%%%%%%%%%%%%%%%%%%%%%%%%%%%%%%%%%%%%%%%
% SECTION 4: DISCUSSION AND CONCLUSION
%%%%%%%%%%%%%%%%%%%%%%%%%%%%%%%%%%%%%%%%%%%%%%%%%%%%%%%%%%%%%%%%
\section{Discussion and Conclusion}

% We have empirically demonstrated a significant failure of rotational symmetry in a leading image-to-3D generative model—a surprising blind spot given that pose invariance is central to the 3D domain. This finding, a clear example of an undesirable inductive bias learned from canonically-oriented web data, directly contributes to the workshop's theme of "Learnability and Generalization."

% The most compelling aspect of our work is the juxtaposition of the problem's severity with the solution's simplicity. A massive, state-of-the-art model fails, yet a tiny CNN pre-processor provides a near-perfect fix. This raises a crucial question for the field, echoing the workshop's prompt: "Should these structures be 'hardwired' in the architectures or learned?" Our results suggest that for fundamental, well-defined symmetries, relying on scale and self-supervision to learn them implicitly is inefficient and unreliable. A more principled, modular approach—where a small, specialized model handles a core task like orientation detection—can lead to far more robust and dependable systems. We need to be smarter in our architectural design, not just bigger.

% Future work should investigate the prevalence of this canonical view bias across other generative models and modalities. Building architectures with better, more explicit reasoning about geometric transformations is a critical step towards creating truly generalizable and reliable generative AI.
\noindent \textbf{Discussion.}  
Our findings reveal a clear canonical-view bias in a state-of-the-art image-to-3D pipeline. Specifically, rotating input images by \(90^\circ\), \(180^\circ\), or \(270^\circ\) consistently leads to lower ULIP similarity scores across object categories such as airplanes, chairs, and cars, with the \(0^\circ\) (canonical) view performing best in all cases (Fig.~\ref{fig:main_result}). Qualitative results further illustrate structured geometric artifacts in outputs from non-canonical views, for example, collapsed or sheared wings and duplicated chair legs, while canonical inputs yield stable and coherent reconstructions (Fig.~\ref{fig:placeholder}). 

Notably, increasing the number of diffusion inference steps fails to mitigate this discrepancy, suggesting that the degradation stems from limitations in the learned representation itself rather than from insufficient sampling or premature convergence. To address this, we demonstrate that a lightweight auxiliary network, trained to predict input rotation, can effectively recover performance to near-canonical levels, without any modification to the underlying generative model (Table~\ref{tab:results}).

\vspace{0.5em}
\noindent \textbf{Limitations and Future Work.}  
We identify three primary limitations of our current study, each of which motivates future research directions:

\begin{enumerate}
    \item \textbf{Discrete Rotation Group (\(C_4\)):}  
Our analysis is limited to 90° in-plane rotations, corresponding to the discrete cyclic group \(C_4\). Extending this evaluation to continuous in-plane rotations (\(SO(2)\)) remains an open direction. Additionally, it would be valuable to explore whether orientation-invariant or equivariant architectures, such as steerable CNNs that produce continuously equivariant features, can mitigate this bias more effectively than discrete classification approaches.

    \item \textbf{Model Scope:}  
    Our experiments focus exclusively on Hunyuan3D~2.0~\cite{zhao2025hunyuan3d}. However, other image-to-3D models such as TripoSR~\cite{TripoSR2024}, OpenLRM~\cite{openlrm}, and others, are also promising candidates for assessing rotation-induced degradation and for testing the generality of our correction strategy.

\item \textbf{Category Coverage:}  
This study focuses on object categories with well-defined canonical views, such as airplanes, chairs, and cars. An open question is whether similar canonical-view biases persist in more diverse or less geometrically constrained categories.

\end{enumerate}

\noindent \textbf{Conclusion.} 
% These findings suggest that merely scaling up models may not be sufficient to overcome such limitations. Instead, it may be fruitful to consider the natural symmetries inherent in the problem and incorporate them into the model architecture. Doing so could lead to more robust, generalizable, and geometrically-aware generative systems.
These findings suggest that scaling alone is unlikely to resolve the observed failure modes. We demonstrate that a lightweight CNN, trained to detect and correct input orientation, can effectively mitigate these issues and restore model performance, all without modifying the generative backbone. While our study focuses on a constrained problem setting, the results highlight the importance of encoding natural symmetries into the representation pipeline, raising interesting questions for future research.

\newpage
\bibliographystyle{plain}
\bibliography{refs}

\newpage
\appendix
\section{Appendix}\label{appendix}

\label{app:orientation_corrector}

\subsection{Model Architecture}

The orientation classifier is based on EfficientNetV2-S \cite{tan2021efficientnetv2smallermodelsfaster}. 

\textbf{Architecture of EfficientNetV2-S}
\begin{itemize}
    \item \textbf{Base Model:} EfficientNetV2-S with ImageNet-1K V1 pretrained weights
    \item \textbf{the Number of Parameters:} 24M 
    \item \textbf{Input Resolution:} 384×384 pixels
    \item \textbf{network architecture:}
    \begin{center}
\begin{tabular}{ |c|c|c|c|c| } 
 \hline
 Stage & Operator & Stride & Channels & Layers \\ 
0  & Conv3x3  & 2  & 24 &  1\\ 
1  & Fused-MBConv1,  k3x3  & 1  & 24  & 2\\ 
2  & Fused-MBConv4,  k3x3  & 2  & 48  & 4\\ 
3  & Fused-MBConv4,  k3x3  & 2  & 64  & 4\\ 
4  & MBConv4, k3x3,    SE0.25 &  2  & 128  & 6\\ 
5  & MBConv6,   k3x3,    SE0.25  & 1  & 160 &  9\\ 
6  & MBConv6,   k3x3,    SE0.25  & 2  & 256  & 15\\ 
7  & Conv1x1 + Pooling + FC  & -  & 1280  & 1\\ 
 \hline
\end{tabular}
\end{center}
    \item \textbf{Feature Dimension:} 1280-dimensional feature vector (output of stage 6, input to classifier)

\end{itemize}

\textbf{From EfficientNetV2-S to Orientation Classifier:}
The Orientation classifier is trained by employing a partial unfreezing approach to balance adaptation and stability. The first 3 feature blocks remain frozen with their ImageNet weights, while the final 5 feature blocks and the classifier head are fine-tuned. The classifier head is replaced with a custom module:Dropout layer (dropout rate 0.3) + Linear layer ($1280 \rightarrow 4 $ classes). The number of parameters of the orientation corrector is around 20.3M .

\subsection{Training Configuration}

\textbf{Dataset:}
The model is trained on a comprehensive dataset of 189,018 unique images from multiple sources:
\begin{itemize}
    \item Microsoft COCO dataset: general object recognition
    \item AI-generated vs. real image datasets: art and illustration orientation
    \item TextOCR dataset: text-oriented image handling
    \item Curated personal photographs: edge cases and unique examples
\end{itemize}

Each image is augmented by generating four rotated versions ($0^\circ, 90^\circ, 180^\circ, 270^\circ$), resulting in 756,072 total training samples. The dataset is split into 604,857 training samples (80\%) and 151,215 validation samples (20\%).

\textbf{Data Augmentation:}
During training, images undergo the following augmentations (applied in order):
\begin{enumerate}
    \item \textbf{RandomResizedCrop:} Scale range (0.85, 1.0), output size 384×384
    \item \textbf{ColorJitter:} Brightness ($\pm$20\%), Contrast ($\pm$20\%), Saturation ($\pm$20\%), Hue ($\pm$10\%)
    \item \textbf{Normalize:} ImageNet statistics (mean: [0.485, 0.456, 0.406], std: [0.229, 0.224, 0.225])
    \item \textbf{RandomErasing:} Probability 0.25, scale (2-10\% of image area)
\end{enumerate}

These augmentations preserve orientation information while introducing variability to improve generalization. Validation images use a simpler pipeline: resize to 416×416, center crop to 384×384.

\end{document}